\documentclass[runningheads]{llncs}
\usepackage[T1]{fontenc}
\usepackage{microtype}
\usepackage[colorlinks,
            citecolor=gray,
            linkcolor=gray,
            urlcolor=gray,
            ]{hyperref}

\usepackage{graphicx}
\usepackage[misc]{ifsym}
\usepackage{amsmath,amssymb,amstext,bm}
\usepackage{orcidlink}
\usepackage{wrapfig}
\usepackage{booktabs}
\usepackage{multirow}

\usepackage{xcolor}
\definecolor{ourblue}{HTML}{1f77b4}
\definecolor{ourred}{HTML}{cd1f1f}

\newcommand{\halfquad}{\:\;}

\newcommand{\R}{\mathbb{R}}
\newcommand{\bbS}{\mathbb{S}}
\newcommand{\f}{\mathbf{f}}

\newcommand{\s}{\mathbf{s}}
\newcommand{\x}{\mathbf{x}}

\newcommand{\z}{\mathbf{z}}

\newcommand{\repeatthanks}{\textsuperscript{\thefootnote}}

\begin{document}
\title{Principled Interpolation in Normalizing Flows}
\author{
Samuel~G.~Fadel (\Letter) \thanks{Equal contribution. \\ This paper is accepted at ECML PKDD 2021.}\inst{1,2}\orcidlink{0000-0002-4459-4336}
\and
Sebastian~Mair\repeatthanks\inst{2}\orcidlink{0000-0003-2949-8781}
\and \\ 
Ricardo~da~S.~Torres\inst{3}\orcidlink{0000-0001-9772-263X}
\and
Ulf~Brefeld\inst{3}\orcidlink{0000-0001-9600-6463}
}
\authorrunning{Fadel et al.}
\toctitle{Principled Interpolation in Normalizing Flows}
\tocauthor{Samuel~G.~Fadel, Sebastian~Mair, Ricardo~da~S.~Torres, Ulf~Brefeld}
\institute{
University of Campinas, Brazil\\
\email{samuel.fadel@ic.unicamp.br}
\and
Leuphana University, Germany\\
\email{\{mair,brefeld\}@leuphana.de}
\and
Norwegian University of Science and Technology, Norway\\
\email{ricardo.torres@ntnu.no}
}
\maketitle
\begin{abstract}
Generative models based on normalizing flows are very successful in modeling complex data distributions using simpler ones.
However, straightforward linear interpolations show unexpected side effects, as interpolation paths lie outside the area where samples are observed.
This is caused by the standard choice of Gaussian base distributions and can be seen in the norms of the interpolated samples as they are outside the data manifold.
This observation suggests that changing the way of interpolating should generally result in better interpolations, but it is not clear how to do that in an unambiguous way.
In this paper, we solve this issue by enforcing a specific manifold and, hence, change the base distribution, to allow for a principled way of interpolation.
Specifically, we use the Dirichlet and von Mises-Fisher base distributions on the probability simplex and the hypersphere, respectively.
Our experimental results show superior performance in terms of bits per dimension, Fréchet Inception Distance (FID), and Kernel Inception Distance (KID) scores for interpolation, while maintaining the generative performance.

\end{abstract}

\section{Introduction}\label{sec:intro}

Learning high-dimensional densities is a common task in unsupervised learning.
Normalizing flows~\cite{tabak2010density,tabak2013family,rippel2013high,dinh2014nice,rezende2015variational} provide a framework for transforming complex distributions into simple ones:
a chain of $L$ parametrized bijective functions $\f=\f_1 \circ \f_2\circ \cdots \circ \f_L$ converts data into another representation that follows a given base distribution.
The likelihood of the data can then be expressed as the likelihood of the base distribution and the determinants of the Jacobians of the transformations $\f_i$.
In contrast to generative adversarial networks (GANs)~\cite{goodfellow2014generative}, the likelihood of the data can be directly optimized, leading to a straightforward training procedure.
Moreover, unlike other approaches, such as variational autoencoders (VAEs)~\cite{kingma2013auto,kingma2019introduction}, there is no reconstruction error since all functions $\f_i$ within this chain are bijections.

\begin{figure}[t]
\center
\includegraphics[width=.75\textwidth]{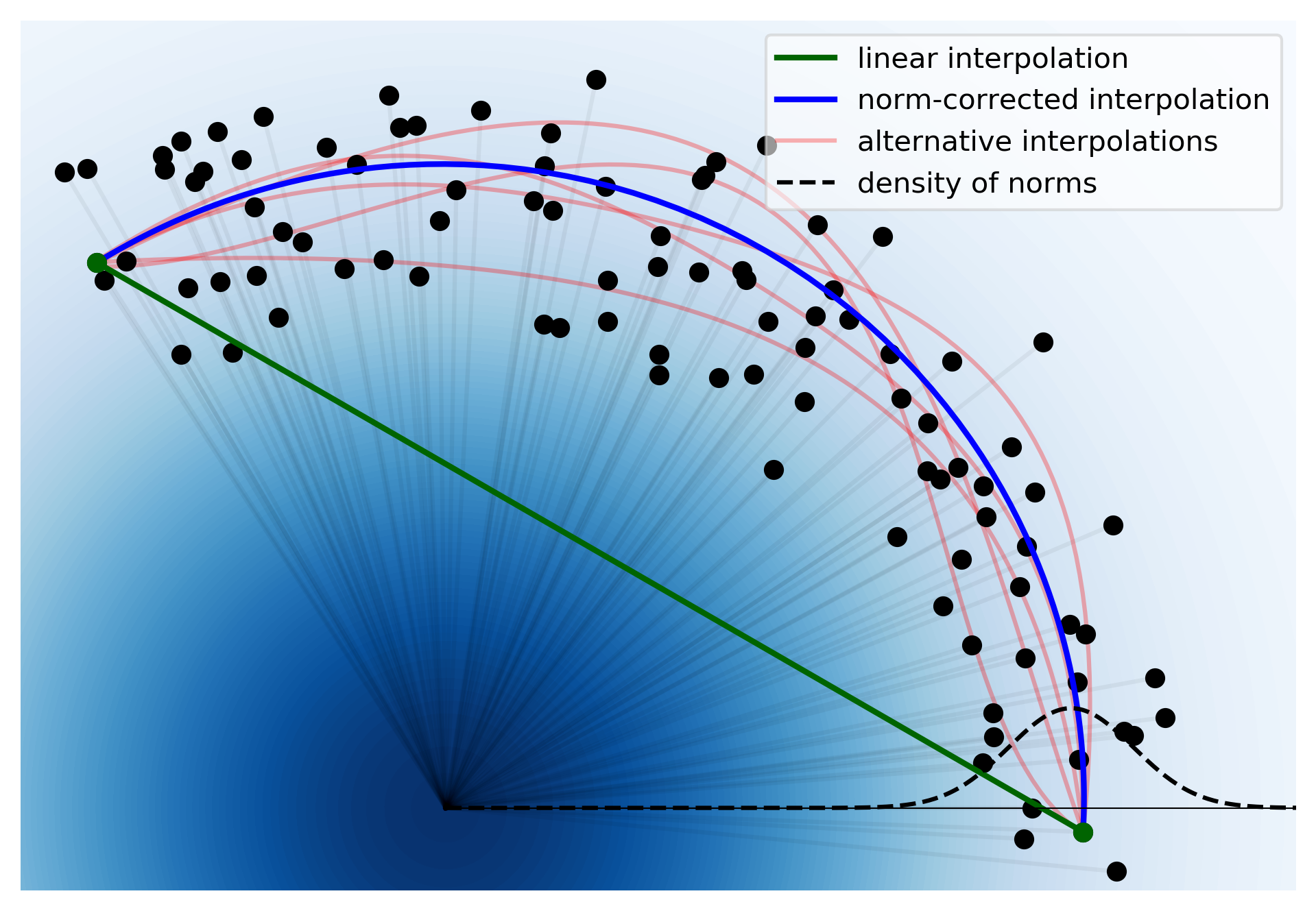}
\caption{Illustration of different interpolation paths of points from a high-dimensional Gaussian. The figure also shows that, in high dimensions, points are not concentrated at the origin.}
\label{fig:intro}
\end{figure}

In flow-based generative models, data are generated by drawing samples from a base distribution, where the latter is usually given by a simple distribution, such as a standard Gaussian~\cite{papamakarios2019normalizing}.
The Gaussian samples are then mapped to real data using the chain ${\bf f}$. 
A prevalent operation is to linearly interpolate samples and consider the interpolation path in data space.
In generative modeling, interpolations are frequently used to evaluate the quality of the learned model and to demonstrate that the model generalizes beyond what was seen in the training data~\cite{radford2015unsupervised}.

The consequences for interpolation, however, are not immediately apparent for Gaussian base distributions.
Figure~\ref{fig:intro} shows a linear interpolation (\emph{lerp}) of high-dimensional samples from a Gaussian.
The squared Euclidean norms of the samples follow a $\chi^2_d$-distribution as indicated by the dashed black line.
Data points have an expected squared Euclidean norm of length $d$, where $d$ is the dimensionality.
This implies that there is almost no point around the origin.
As seen in the figure, the norms of a linear interpolation path (green line) of two samples drop significantly and lie in a low-density area w.r.t. the distribution of the norms (dashed black line)~\cite{white2016sampling}.

Instead of a linear interpolation (green line), an interpolation that preserves the norm distribution of interpolants is clearly preferable (blue and red lines):
the blue and red interpolation paths stay in the data manifold and do not enter low-density areas.
The observation suggests that interpolated samples with norms in a specific range should generally result in better interpolations.
This can be achieved, i.e., by shrinking the variance of the density or norms (dashed lines), which yields a subspace or manifold with a fixed norm.

In this paper, we propose a framework that respects the norm of the samples and allows for a principled interpolation, addressing the issues mentioned above.
We study base distributions on supports that have a fixed norm.
Specifically, we consider unit $p$-norm spheres for $p\in\{1,2\}$, leading to the Dirichlet ($p=1$) and 
the von Mises-Fisher ($p=2$) distributions, respectively.
The conceptual change naturally implies technical difficulties that arise with restricting the support of the base distribution to the simplex or the unit hypersphere. 
We thus need to identify appropriate bijective transformations into unit $p$-norm spheres.

The next sections are organized as follows.
In Section~\ref{sec:heuristic}, we propose a simple heuristic to the problem and discuss its problems before we introduce normalizing flows in Section~\ref{sec:normalizingflows}.
Section~\ref{sec:framework} contains the main contribution, a framework for normalizing flows onto unit $p$-norm spheres.
Empirical results are presented in Section~\ref{sec:experiments}, and related work is discussed in Section~\ref{sec:related_work}.
Section~\ref{sec:conclusion} provides our conclusions.

\section{An Intuitive Solution}\label{sec:heuristic}

The blue path in Figure~\ref{fig:intro} is obtained by a norm correction of the linear interpolation via also interpolating the norms.
Mathematically, that is
\begin{align}\label{eq-heuristic}
\gamma(\lambda) = \underbrace{\left ( (1-\lambda) \z_a + \lambda \z_b \right )}_{\text{linear interpolation}} \cdot \underbrace{\frac{(1-\lambda) \|\z_a\|_2 + \lambda \|\z_b\|_2}{\|(1-\lambda) \z_a + \lambda \z_b\|_2}}_{\text{norm correction}},
\end{align}
for endpoints $\z_a, \z_b$ and $\lambda \in [0,1]$.
We refer to this approach as norm-corrected linear interpolation (\emph{nclerp}).
However, the depicted red lines also stay within the manifold, hence it remains unclear how a unique interpolation path can be obtained.

\begin{figure}[t]
\center
\includegraphics[width=\textwidth]{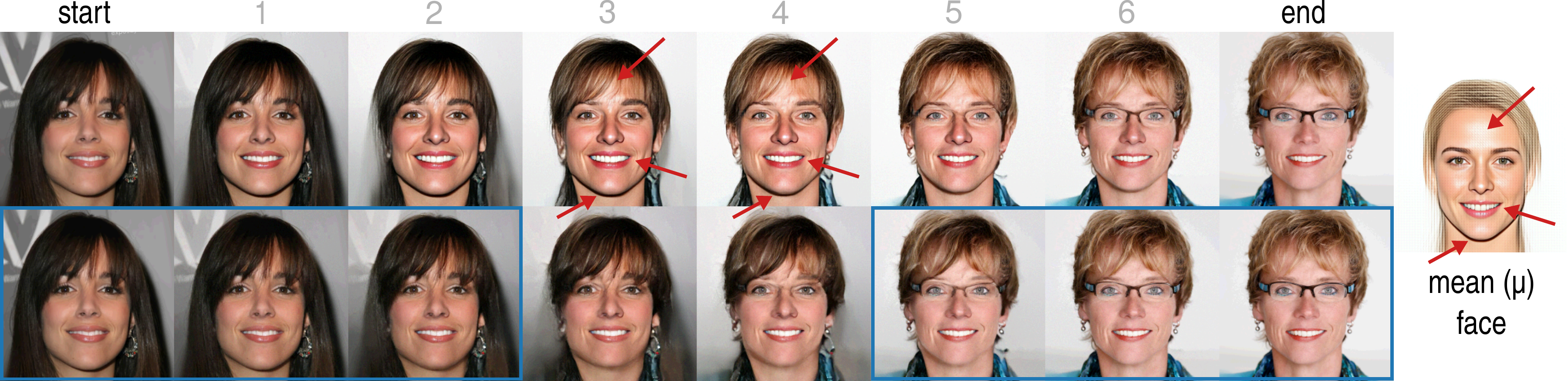}
\caption{Interpolation of samples from CelebA. \emph{Top}: a linear interpolation path. The central images resemble features of the mean face as annotated in \color{ourred}red\color{black}. \emph{Bottom}: an alternative interpolation path using a norm-correction. Note that the first and last three images are almost identical as annotated in \color{ourblue}blue\color{black}. \emph{Right}: decoded expectation of base distribution, i.e., the mean face.}
\label{fig:interpol1}
\end{figure}

Figure~\ref{fig:interpol1} depicts two interpolation paths for faces taken from CelebA~\cite{karras2018progressive} created using Glow~\cite{kingma2018glow}, a state-of-the-art flow-based model that uses a standard Gaussian as base distribution.
The leftmost and rightmost faces of the paths are real data, while the other ones are computed interpolants.
The face on the right depicts the so-called \emph{mean face}, which is given by the mean of the Gaussian base distribution and is trivially computed by decoding the origin of the space.
The top row shows a linear interpolation similar to the green line in Figure~\ref{fig:intro}.
The interpolation path is close to the origin, and the interpolants consequentially resemble features of the mean face, such as the nose, mouth, chin, and forehead shine, which neither of the women have.
We highlighted those features in red in Figure~\ref{fig:interpol1}.

By contrast, the bottom row of Figure~\ref{fig:interpol1} shows the norm-corrected interpolation sequence (as the blue line in Figure~\ref{fig:intro}):
the background transition is smooth and not affected by the white of the mean face, and also subtleties like the shadow of the chin in the left face smoothly disappears in the transition.
The norm correction clearly leads to a better transition from one image to the other. 
However, the simple heuristic in Equation~\eqref{eq-heuristic} causes another problem:
the path after norm correction is no longer equally spaced when $\lambda$ values are equally spaced in $[0, 1]$.
Implications of this can be seen in blue in the bottom row of Figure~\ref{fig:interpol1}, where the first and last three faces are almost identical.
We provide additional examples in the supplementary material.

\begin{figure}[t]
\center
\includegraphics[width=\textwidth]{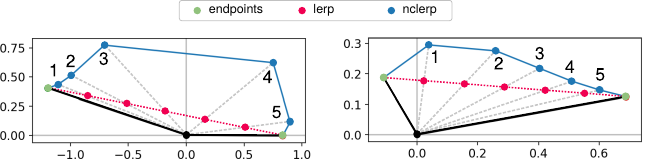}
\caption{Two examples showing the issues caused by a norm-corrected linear interpolation (\emph{nclerp}).}
\label{fig:lerp_slerp}
\end{figure}

In Figure~\ref{fig:lerp_slerp}, we illustrate two examples, comparing a linear interpolation (\emph{lerp}) and a norm-corrected linear interpolation (\emph{nclerp}) between points from a high-dimensional Gaussian (green points).
For equally-spaced $\lambda$ values in $[0,1]$, a linear interpolation yields an equally-spaced interpolation path (red line).
Evidently, the norm-corrected interpolation (blue line) keeps the norms of interpolants within the range observed in data.

However, the interpolants are no longer evenly spaced along the interpolation path.
Hence, control over the interpolation mixing is lost.
This problem is more pronounced on the left example, where points 1 through 3 are closer to the start point, while points 4 and 5 are closer to the endpoint.
Consequently, evaluations such as Fr\'echet Inception Distance (FID) scores~\cite{heusel2017gans} will be affected.
Such scores are computed by comparing two sets of samples, in this case, the real data and interpolated data.
As those points will be clearly more similar to the endpoints, which are samples from the data set itself, the scores are in favor of the norm-corrected interpolation.

\section{Normalizing Flows}\label{sec:normalizingflows}

Let $\mathcal{X} = \{\x_1, \ldots, \x_n \} \subset \R^d$ be instances drawn from an unknown distribution $p_x(\x)$.
The goal is to learn an accurate model of $p_x(\x)$.
Let $\f^{(\theta)} : \R^d \to \R^d$ be a bijective function parametrized by $\theta$.
Introducing $\z = \f^{(\theta)}(\x)$ and using the change of variable theorem allows us to express the unknown $p_x(\x)$ by a (simpler) distribution $p_z(\z)$, defined on $\z \in \R^d$, given by
\begin{align*}
p_x(\x) = p_z\left(\f^{(\theta)}(\x)\right) \left|\det J_\f^{(\theta)}(\x)\right|,
\end{align*}
where $J_\f^{(\theta)}(\x)$ is the Jacobian matrix of the bijective transformation $\f^{(\theta)}$.
We denote $p_z(\z)$ as the base distribution and drop the subscript of the distribution $p$ whenever it is clear from the context.

\begin{figure}[t]
\center
\includegraphics[width=\textwidth]{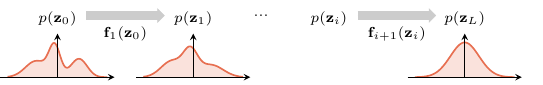}
\caption{An example of a one-dimensional normalizing flow.}
\label{fig:normalizingflow}
\end{figure}

Representing $\f^{(\theta)}$ as a chain of $L$ parametrized bijective functions, i.e., $\f^{(\theta)} = \f_L^{(\theta)} \circ \f_{L-1}^{(\theta)} \circ \cdots \circ \f_1^{(\theta)}$, creates a \textit{normalizing flow} that maps observations $\x$ into representations $\z$ that are governed by the base distribution $p(\z)$.
Let $\z_0 = \x$ be the input data point and $\z_L = \z$ be the corresponding output of the chain, where every intermediate variable is given by $\z_i = \f_i(\z_{i-1})$, where $i=1,\ldots,L$.
A one-dimensional example is depicted in Figure~\ref{fig:normalizingflow}.
The data log-likelihood can then be expressed as the log-likelihood of the base distribution and the log-determinant of the Jacobians of each transformation as
\begin{align*}
\log p(\x) = \log p(\z) + \sum_{i=1}^L \log \left| \det J_{\f^{(\theta)}_i}(\z_{i-1}) \right|.
\end{align*}
Flow-based generative models can be categorized by how the Jacobian structure of each transformation $\f_i$ is designed since computing its determinant is crucial to its computational efficiency.
The Jacobians either have a lower triangular structure, such as autoregressive flows~\cite{kingma2016improved}, or a structured sparsity, such as coupling layers in RealNVP~\cite{dinh2016density} and Glow~\cite{kingma2018glow}.
Transformations with free-form Jacobians allow much higher expressibility by replacing the computation of the determinant with another estimator for the log-density \cite{chen2019residual}.
For more information regarding flow-based generative models, we refer the reader to \cite{papamakarios2019normalizing}.

In the remainder, we simplify the notation by dropping the superscript $\theta$ from $\f$.
We also note that a normalizing flow defines a generative process.
To create a new sample $\x$, we first sample $\z$ from the base distribution $p(\z)$ and then transform $\z$ into $\x$ using the inverse chain of transformations $\f^{-1}$.

\section{Base Distributions on p-Norm Spheres}\label{sec:framework}

Motivated by earlier observations illustrated in Figures~\ref{fig:intro} and \ref{fig:interpol1}, we intend to reduce ambiguity by shrinking the variance of the norms of data.
We achieve this by considering base distributions on restricted subspaces.
More specifically, we focus on unit $p$-norm spheres defined by
\begin{align}\label{eq:p_unit_norm}
\mathbb{S}_p^d = \left\{ \z \in \R^{d+1} \ \Bigg | \ \|\z\|_p^p = \sum_{j=1}^{d+1} |z_j|^p = 1 \right\}.
\end{align}
We distinguish two choices of $p$ and discuss the challenges and desirable properties that ensue from their use.
We consider $p \in \{1, 2\}$ as those allow us to use well-known distributions, namely the Dirichlet distribution for $p=1$ and the von Mises-Fisher distribution for $p=2$.

\subsection{The Case p=1}

For $p = 1$, the Dirichlet distribution defined on the standard simplex $\Delta^d$ is a natural candidate.
Its probability density function is given by
\begin{align*}
p(\s) &= \frac{1}{Z(\alpha)} \prod_{k=1}^{d+1} s_k^{\alpha_k-1},\\
\text{with} \quad Z(\alpha) &= \frac{\prod_{k=1}^{d+1} \Gamma\left(\alpha_k\right)}{\Gamma\left(\sum_{k=1}^{d+1} \alpha_k\right)},
\end{align*}
where $\Gamma$ is the gamma function and $\alpha_k>0$ are the parameters.
In order to make use of it, we also need to impose a non-negativity constraint in addition to Equation~\eqref{eq:p_unit_norm}.

Let $\z \in \R^d$ be an unconstrained variable.
The function $\phi : \R^d \to (0, 1)^d$ transforms $\z$ into a representation $\s$ by first transforming each dimension $z_k$ into intermediate values $v_k$ with
\begin{align*}
v_k = \sigma\left(z_k -\log\left(d+1-k\right)\right)
\end{align*}
which are used to write $\s$ as
\begin{align*}
s_k = \left(1-\sum_{l=1}^{k-1} s_l\right) \cdot v_k,
\end{align*}
where $\sigma(\cdot)$ denotes the sigmoid function.
We note a few details of this transformation.
First, a property of $\phi$ is that $0 < \sum_{k=1}^d s_k < 1$.
Therefore, a point in $\Delta^d$ can be obtained with an implicit additional coordinate $s_{d+1} = 1 - \sum_{k=1}^{d} s_k$.
Second, the difference in dimensionality does not pose a problem for computing its Jacobian as $\phi$ establishes a bijection within $\R^d$ while the mapping to $\Delta^d$ is given implicitly.
Third, $\phi$ maps $\z=\mathbf{0}$ to the center of the simplex $\s=(d+1)^{-1}\mathbf{1}$.
Fourth, since $\s$ consists of solely positive numbers which sum up to one, numerical problems may arise for high-dimensional settings.
We elaborate on this issue in Section~\ref{sec:experiments}.

The Jacobian $J_\phi$ has a lower triangular structure and solely consists of non-negative entries.
Hence, the log-determinant of this transformation can be efficiently computed in $\mathcal{O}(d)$ time via
\begin{align*}
\log | \det J_\phi |
= \sum_{k=1}^{d} &\log \left(v_k \left(1-v_k\right)\right) \\
               + &\log \left(1-\sum_{l=1}^{k-1} s_l \right).
\end{align*}
The inverse transformation $\phi^{-1} : \R^d \to \R^d$ is given by
\begin{align*}
z_k &= \sigma^{-1}\left(\frac{s_k}{1-\sum_{l=1}^{k-1} s_l}\right) + \log(d+1-k).
\end{align*}

The interpolation of two points $\mathbf{a}, \mathbf{b} \in \Delta^d$ within the unit simplex is straightforward.
A linear interpolation $(1-\lambda) \mathbf{a} + \lambda \mathbf{b}$ using $\lambda \in [0,1]$ is guaranteed to stay within the simplex by definition.

\subsection{The Case p=2}\label{sec:vmf}

For $p = 2$, data points lie on the surface of a $d$-dimensional hypersphere.
The von Mises-Fisher (vMF) distribution, defined on $\bbS_2^d$, is frequently used in directional statistics.
It is parameterized by a \emph{mean direction} $\bm{\mu} \in \bbS_2^d$ and a \emph{concentration} $\kappa \geq 0$, with a probability density function given by
\begin{align*}
p(\s) &= C_{d+1}(\kappa) \exp(\kappa \bm{\mu}^\top \s),\\
\text{with} \quad C_\nu(\kappa) &= \frac{\kappa^{\nu/2-1}}{(2 \pi)^{\nu/2}I_{\nu/2-1}(\kappa)},
\end{align*}
where $I_{w}$ denotes the modified Bessel function of the first kind at order $w$.

Again, let $\z \in \R^d$ be an unconstrained variable.
We employ a stereographic projection, for both its invertibility and its Jacobian, whose log-determinant can be efficiently computed.
The transformation $\psi:\R^d \to \bbS_2^d$ maps a point $\z \in \R^d$ to a point $\s \in \bbS_2^d \subset \R^{d+1}$ on the hypersphere via
\begin{align*}
\psi(\z) = \s &= \left[ \begin{matrix} \z \rho_\z \\ 1 - \rho_\z \end{matrix} \right],
  \quad \text{with} \quad \rho_\z = \frac{2}{1 + \|\z\|^2}.
\end{align*}
\begin{wrapfigure}{r}{0.4\textwidth}
\center
\includegraphics[width=0.33\textwidth]{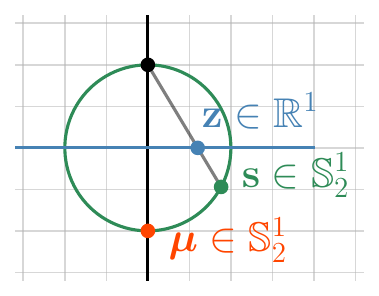}
\caption{A stereographic projection mapping $\z \in \R^1$ to $\s \in \bbS^1_2$ using the north pole depicted as a black dot. The mean direction $\bm{\mu} \in \bbS^1_2$ is shown in orange.}
\label{fig:map}
\end{wrapfigure}
The transformation $\psi$, which has no additional parameters, ensures that its image is on the unit hypersphere, allowing the use of a vMF distribution to model $p(\s)$.
Two points in $\bbS_2^d$ are of special interest, namely the \textit{south pole} and the \textit{north pole}, where the last coordinate of $\s$ is either~$-1$ or~$1$, respectively.
By construction, the transformation is symmetric around zero and sends $\z=\mathbf{0}$ to the \textit{south pole}, which we choose as the mean direction $\bm{\mu}$.
Furthermore, it is bijective up to an open neighborhood around the \textit{north pole}, as $\rho_\z \to 0$ whenever $\|\z\|^2 \to \infty$.
For this reason, we avoid choosing a uniform distribution on the hypersphere, which is obtained for $\kappa = 0$.
Figure~\ref{fig:map} shows an example.

Contrary to the previous case, the log-determinant of $J_\psi$ alone is not enough to accommodate the density change when transforming from $\R^d$ to $\bbS^d_2$~\cite{gemici2016normalizing}.
The correct density ratio change is scaled by $\sqrt{\det J_\psi^\top J_\psi}$ instead, whose logarithm can be computed in $\mathcal{O}(d)$ time as
\begin{align*}
\log \sqrt{\det J_\psi^\top(\z) J_\psi(\z)} = d \log \frac{2}{1 + \|\z\|^2} = d \log \rho_\z,
\end{align*}
with $\rho_\z$ given as stated above.
The inverse function $\psi^{-1} : \bbS_2^d \subset \R^{d+1} \to \R^d$ is
\begin{align*}
\psi^{-1}(\s) = \z = \frac{[\s]_{1:d}}{1 - [\s]_{d + 1}},
\end{align*}
where $[\s]_{1:d}$ denotes the first $d$ coordinates of $\s$ and $[\s]_{d+1}$ is the $(d+1)$-th coordinate of $\s$.

To interpolate points on the hypersphere, a spherical linear interpolation (\emph{slerp})~\cite{shoemake1985animating} can be utilized.
It is defined as follows.
Let $\s_a$ and $\s_b$ be two unit vectors and $\omega=\cos^{-1}(\s_a^\top\s_b)$ be the angle between them.
The interpolation path is then given by
\begin{align*}
\gamma(\lambda) = \frac{\sin((1-\lambda)\omega)}{\sin(\omega)} \s_a + \frac{\sin(\lambda\omega)}{\sin(\omega)} \s_b,
\halfquad \text{for} \halfquad \lambda \in [0,1].
\end{align*}

\section{Experiments}\label{sec:experiments}

We now evaluate the restriction of a normalizing flow to a unit $p$-norm sphere and compare them to a Gaussian base distribution.
As we focus on a principled way of interpolating in flow-based generative models, we employ a fixed architecture per data set instead of aiming to achieve state-of-the-art density estimation.
We use Glow~\cite{kingma2018glow} as the flow architecture for the experiments in the remainder of this section.
However, our approach is not limited to Glow, and the transformations and changes in the base distribution can also be used in other architectures.
We also do not compare against other architectures as our contribution is a change of the base distribution, allowing for better interpolations.

\subsection{Performance Metrics and Setup}

Performance is measured in terms of \emph{bits per dimension (BPD)}, calculated using $\log_2 p(\x)$ divided by $d$; \emph{Fr\'echet Inception Distance (FID) scores}, which have been shown to correlate highly with human judgment of visual quality~\cite{heusel2017gans}; and \emph{Kernel Inception Distance (KID) scores}~\cite{binkowski2018demystifying}.
KID is similar to FID as it is based on Inception scores~\cite{salimans2016improved}.
While the FID first fits a Gaussian distribution on the scores of a reference set and a set of interest and then compares the two distributions, the KID score is non-parametric, i.e., it does not assume any distribution and compares the Inception scores based on Maximum Mean Discrepancy (MMD).
We follow previous work \cite{binkowski2018demystifying} and employ a polynomial kernel with degree three for our evaluations.

We measure bits per dimension on the test set and on interpolated samples.
FID and KID scores are evaluated on generated and interpolated samples and then compared to a reference set, which is the training data.
When generating data, we draw as many samples from the base distribution as we have for training.
For interpolation, we focus on interpolation within classes and adopt regular linear interpolation for Gaussian-distributed samples, while using a spherical linear interpolation on the sphere for vMF-distributed samples.
In this operation, we sample $n/5$ pairs of images from the training set and generate five equally spaced interpolated data instances per pair, resulting in $n$ new images.
From those interpolation paths, we only use the generated points and not the points which are part of training data.
Hence, we are only considering previously unseen data.
\looseness=-1

We also compare against the norm-corrected linear interpolation (\emph{nclerp}) defined in Equation~\eqref{eq-heuristic}.
Note that a linearly spaced interpolation path is no longer linearly spaced after norm correction.
The resulting interpolation paths are composed of images located closer to the endpoints and thus bias the evaluation.
We include the results nevertheless for completeness and refer to the supplementary material for a detailed discussion.

The reported metrics are averages over three independent runs and include standard errors.
The code is written in PyTorch~\cite{PyTorch2019}. 
All experiments run on an Intel Xeon CPU with 256GB of RAM using an NVIDIA V100 GPU.

\subsection{Data}

In our experiments, we utilize \textbf{MNIST}~\cite{lecun2010mnist}, \textbf{Kuzushiji-MNIST}~\cite{clanuwat2018deep}, and \textbf{Fashion-MNIST}~\cite{xiao2017fashion}, which contain gray-scale images of handwritten digits, Hiragana symbols, and images of Zalando articles, respectively.
All MNIST data sets consist of 60,000 training and 10,000 test images of size $28\times28$.
In addition, we evaluate on \textbf{CIFAR10}~\cite{krizhevsky2009learning}, which contains natural images from ten classes.
The data set has 50,000 training and 10,000 test images of size $32\times32$.

\begin{table}[t]
\center
\caption{Results for generative modeling averaged over three independent runs including standard errors.}
\label{tab:results_generative}
\begin{tabular}{rcccc}
\toprule
& & Test & \multicolumn{2}{c}{Sample} \\
\cmidrule(l){3-3} \cmidrule(l){4-5}  
& Base dist.           & BPD & FID & KID  \\
\midrule 
\multirow{5}{*}{\rotatebox[origin=c]{90}{MNIST}}
& Gaussian             &         1.59 \tiny{$\pm$   0.06} & \textbf{34.53}\tiny{$\pm$   0.83} & \textbf{0.033}\tiny{$\pm$  0.001} \\
& vMF $\kappa=1d$      & \textbf{1.46}\tiny{$\pm$   0.07} &         40.07 \tiny{$\pm$   2.46} &         0.037 \tiny{$\pm$  0.001} \\
& vMF $\kappa=1.5d$    &         1.54 \tiny{$\pm$   0.09} &         40.39 \tiny{$\pm$   1.40} &         0.036 \tiny{$\pm$  0.001} \\
& vMF $\kappa=2d$      &         1.82 \tiny{$\pm$   0.08} &         39.82 \tiny{$\pm$   0.26} &         0.038 \tiny{$\pm$  0.001} \\
& Dirichlet $\alpha=2$ &         1.76 \tiny{$\pm$   0.12} &         40.08 \tiny{$\pm$   0.72} &         0.039 \tiny{$\pm$  0.001} \\
\midrule 
\multirow{5}{*}{\rotatebox[origin=c]{90}{K-MNIST}}
& Gaussian             &         2.58 \tiny{$\pm$   0.11} &         35.34 \tiny{$\pm$   0.76} &         0.041 \tiny{$\pm$  0.001} \\
& vMF $\kappa=1d$      &         2.63 \tiny{$\pm$   0.06} &         36.63 \tiny{$\pm$   0.37} &         0.041 \tiny{$\pm$  0.001} \\
& vMF $\kappa=1.5d$    & \textbf{2.48}\tiny{$\pm$   0.06} & \textbf{35.00}\tiny{$\pm$   0.61} & \textbf{0.040}\tiny{$\pm$  0.001} \\
& vMF $\kappa=2d$      &         2.51 \tiny{$\pm$   0.04} &         36.45 \tiny{$\pm$   0.42} &         0.041 \tiny{$\pm$  0.001} \\
& Dirichlet $\alpha=2$ &         2.50 \tiny{$\pm$   0.05} &         35.54 \tiny{$\pm$   0.39} & \textbf{0.040}\tiny{$\pm$  0.001} \\
\midrule 
\multirow{4}{*}{\rotatebox[origin=c]{90}{F-MNIST}}
& Gaussian             &         3.24 \tiny{$\pm$   0.04} &         66.64 \tiny{$\pm$   1.29} &         0.064 \tiny{$\pm$  0.003} \\
& vMF $\kappa=1d$      & \textbf{3.16}\tiny{$\pm$   0.03} & \textbf{60.45}\tiny{$\pm$   3.34} & \textbf{0.055}\tiny{$\pm$  0.005} \\
& vMF $\kappa=1.5d$    &         3.30 \tiny{$\pm$   0.07} &         61.89 \tiny{$\pm$   1.29} &         0.056 \tiny{$\pm$  0.002} \\
& vMF $\kappa=2d$      &         3.22 \tiny{$\pm$   0.06} &         60.60 \tiny{$\pm$   3.47} & \textbf{0.055}\tiny{$\pm$  0.004} \\
\midrule 
\multirow{4}{*}{\rotatebox[origin=c]{90}{CIFAR10}}
& Gaussian             &         3.52 \tiny{$\pm$   0.01} &         71.34 \tiny{$\pm$   0.45} & \textbf{0.066}\tiny{$\pm$  0.001} \\
& vMF $\kappa=1d$      &         3.43 \tiny{$\pm$   0.00} &         71.07 \tiny{$\pm$   0.78} &         0.069 \tiny{$\pm$  0.001} \\
& vMF $\kappa=1.5d$    & \textbf{3.42}\tiny{$\pm$   0.00} & \textbf{70.58}\tiny{$\pm$   0.40} &         0.068 \tiny{$\pm$  0.001} \\
& vMF $\kappa=2d$      & \textbf{3.42}\tiny{$\pm$   0.01} &         71.00 \tiny{$\pm$   0.28} &         0.068 \tiny{$\pm$  0.001} \\
\bottomrule
\end{tabular}
\end{table}

\subsection{Architecture}

We employ the Adam optimizer~\cite{kingma2014adam} with a learning rate of $10^{-3}$, clip gradients at 50, and use linear learning rate warm-up for the first ten epochs.
Models were trained on MNIST data and CIFAR10 using mini-batches of size 256 and 128, respectively.
All models are trained for 100 epochs without early stopping.
We keep all architectures as close as possible to Glow, with the following deviations.
For MNIST data, we use random channel permutations instead of invertible $1\times1$ convolutions.
The number of filters in the convolutions of the affine coupling layers is 128.
In Glow terms, we employ $L = 2$ levels of $K = 16$ steps each.
For CIFAR10, our models have $L = 3$ levels of $K = 24$ steps each, while the affine coupling layers have convolutions with 512 filters.
The architecture is kept the same across base distributions, except for the additional parameterless transformations to the restricted subspaces introduced in Section~\ref{sec:framework}.

When comparing base distributions, we consider the following hyperparameters.
For the vMF distribution, we use concentration values for which the partition function is finite.
For consistency, the values we use are the same multiples of the data dimensionality $d$ for each data set.
The concentration values for the Dirichlet distribution are set to $\alpha = 2$, which refers to $2 \cdot \mathbf{1}^{d+1} \in \R^{d+1}$.

\begin{table}[t]
\center
\caption{Results for interpolation averaged over three independent runs including standard errors. Interpolations are in-class only and use five intermediate points; \emph{lerp} refers to a linear interpolation; \emph{nclerp} refers to the norm-corrected linear interpolation (Section~\ref{sec:intro}) and \emph{slerp} refers to the spherical interpolation.}
\label{tab:results_interpolation}
\begin{tabular}{rcrccc}
\toprule
& Base dist.           & Type  & BPD & FID & KID  \\
\midrule 
\multirow{6}{*}{\rotatebox[origin=c]{90}{MNIST}}
& Gaussian             & lerp  &         1.33 \tiny{$\pm$   0.05} &         5.10 \tiny{$\pm$   0.14} &         0.003 \tiny{$\pm$  0.000} \\
& Gaussian             & nclerp&         1.44 \tiny{$\pm$   0.06} &         5.12 \tiny{$\pm$   0.30} &         0.003 \tiny{$\pm$  0.000} \\
& vMF $\kappa=1d$      & slerp & \textbf{1.31}\tiny{$\pm$   0.09} & \textbf{3.84}\tiny{$\pm$   0.36} & \textbf{0.002} \tiny{$\pm$  0.000} \\
& vMF $\kappa=1.5d$    & slerp &         1.40 \tiny{$\pm$   0.10} &         4.22 \tiny{$\pm$   0.12} & \textbf{0.002} \tiny{$\pm$  0.000} \\
& vMF $\kappa=2d$      & slerp &         1.63 \tiny{$\pm$   0.10} &         4.45 \tiny{$\pm$   0.06} & \textbf{0.002} \tiny{$\pm$  0.000} \\
& Dirichlet $\alpha=2$ & lerp  &         1.61 \tiny{$\pm$   0.10} &         5.81 \tiny{$\pm$   0.36} &         0.004 \tiny{$\pm$  0.001} \\
\midrule 
\multirow{6}{*}{\rotatebox[origin=c]{90}{K-MNIST}}
& Gaussian             & lerp  &         1.91 \tiny{$\pm$   0.17} &         19.71 \tiny{$\pm$   1.59} &         0.021 \tiny{$\pm$  0.002} \\
& Gaussian             & nclerp&         2.15 \tiny{$\pm$   0.15} &         17.60 \tiny{$\pm$   1.48} &         0.020 \tiny{$\pm$  0.002} \\
& vMF $\kappa=1d$      & slerp &         2.08 \tiny{$\pm$   0.15} &         17.93 \tiny{$\pm$   3.72} &         0.020 \tiny{$\pm$  0.004} \\
& vMF $\kappa=1.5d$    & slerp & \textbf{1.80}\tiny{$\pm$   0.07} &         22.72 \tiny{$\pm$   2.65} &         0.025 \tiny{$\pm$  0.003} \\
& vMF $\kappa=2d$      & slerp &         2.03 \tiny{$\pm$   0.14} & \textbf{14.54}\tiny{$\pm$   2.51} & \textbf{0.016}\tiny{$\pm$  0.003} \\
& Dirichlet $\alpha=2$ & lerp  &         1.81 \tiny{$\pm$   0.04} &         24.09 \tiny{$\pm$   2.35} &         0.026 \tiny{$\pm$  0.003} \\
\midrule 
\multirow{5}{*}{\rotatebox[origin=c]{90}{F-MNIST}}
& Gaussian             & lerp  &         2.84 \tiny{$\pm$   0.10} &         13.06 \tiny{$\pm$   0.62} &         0.007 \tiny{$\pm$  0.001} \\
& Gaussian             & nclerp&         2.93 \tiny{$\pm$   0.03} & \textit{ 7.80}\tiny{$\pm$   0.13} &         0.004 \tiny{$\pm$  0.000} \\
& vMF $\kappa=1d$      & slerp & \textbf{2.66}\tiny{$\pm$   0.03} & \textbf{12.16}\tiny{$\pm$   0.13} & \textbf{0.006}\tiny{$\pm$  0.000} \\
& vMF $\kappa=1.5d$    & slerp &         2.84 \tiny{$\pm$   0.07} &         12.19 \tiny{$\pm$   1.07} & \textbf{0.006}\tiny{$\pm$  0.001} \\
& vMF $\kappa=2d$      & slerp &         2.70 \tiny{$\pm$   0.05} &         15.11 \tiny{$\pm$   0.85} &         0.008 \tiny{$\pm$  0.001} \\
\midrule 
\multirow{5}{*}{\rotatebox[origin=c]{90}{CIFAR10}}
& Gaussian             & lerp  &         2.64 \tiny{$\pm$   0.06} &         58.63 \tiny{$\pm$   1.26} &         0.053 \tiny{$\pm$  0.001} \\
& Gaussian             & nclerp&         3.32 \tiny{$\pm$   0.01} & \textit{14.29}\tiny{$\pm$   0.16} &         0.010 \tiny{$\pm$  0.000} \\
& vMF $\kappa=1d$      & slerp &         2.78 \tiny{$\pm$   0.05} & \textbf{51.08}\tiny{$\pm$   0.37} &         0.010 \tiny{$\pm$  0.000} \\
& vMF $\kappa=1.5d$    & slerp &         2.66 \tiny{$\pm$   0.05} &         55.23 \tiny{$\pm$   5.14} &         0.047 \tiny{$\pm$  0.005} \\
& vMF $\kappa=2d$      & slerp & \textbf{2.58}\tiny{$\pm$   0.08} &         52.65 \tiny{$\pm$   3.34} &         0.044 \tiny{$\pm$  0.004} \\
\bottomrule
\end{tabular}
\end{table}

\begin{figure}[t]
\center
\includegraphics[width=.6\textwidth]{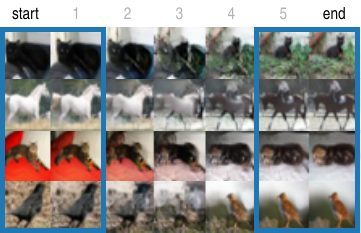}
\caption{Five interpolation paths of the norm-corrected linear interpolation (\emph{nclerp}) depicting the problem of almost repeated endpoints (highlighted in \color{ourblue}blue\color{black}) and thus a biased evaluation on CIFAR10.
}
\label{fig:interpolation_nclerp}
\end{figure}

\subsection{Quantitative Results}

We first evaluate the generative modeling aspects of all competitors.
Table~\ref{tab:results_generative} summarizes the results in terms of bits per dimension on test data and FID and KID scores on generated samples for all data sets.
Experiments with the Dirichlet base distribution were not successful on all data sets.
The restrictions imposed to enable the use of the distribution demand a high numerical precision since every image on the simplex is represented as a non-negative vector that sums up to one.
Consequently, we only report results on MNIST and Kuzushiji-MNIST.
Using the vMF as a base distribution clearly outperforms the Gaussian in terms of bits per dimension on test data.
As seen in the FID and KID scores, we perform competitive compared to the Gaussian for generating new data.
Hence, the generative aspects of the proposed approach are either better or on par with the default choice of a Gaussian.
Note that lower bits per dimension on test data and lower FID/KID scores on generated data might be obtained with more sophisticated models.

We now evaluate the quality of interpolation paths generated via various approaches.
Table~\ref{tab:results_interpolation} shows the results in terms of bits per dimension, FID, and KID scores for all data sets.
The experiments confirm our hypothesis that an interpolation on a fixed-norm space yields better results as measured in bits per dimension, FID, and KID scores.
The norm-corrected interpolation yields better FID and KID scores for Fashion-MNIST and CIFAR10.
However, this heuristic produces interpolation paths that are biased towards the endpoints \mbox{(cf. Figure~\ref{fig:lerp_slerp})} and hence are naturally closer to observed data, thus yield better FID and KID scores.
This is depicted in Figure~\ref{fig:interpolation_nclerp} where the first and last interpolant is very close to real data.
More results on general interpolations within classes and across classes are provided in the supplementary material.

\begin{figure}[h!]
\center
\includegraphics[width=\textwidth]{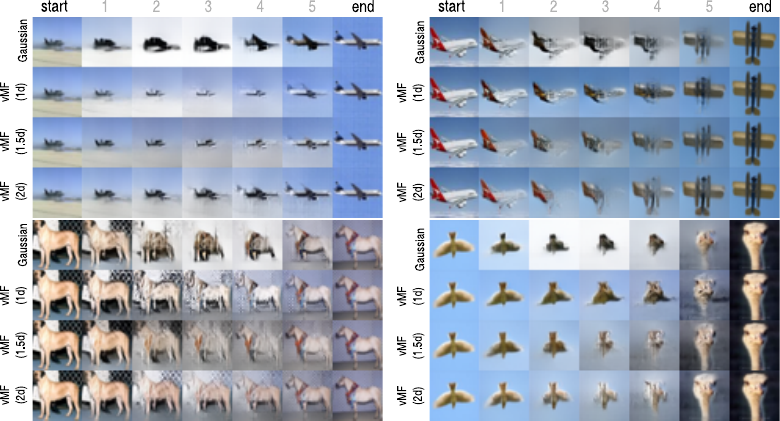}
\caption{Interpolation paths of four pairs of data from CIFAR10 using different models.}
\label{fig:interpolation_cifar10}
\end{figure}

\subsection{Qualitative Results}

Figure~\ref{fig:interpolation_cifar10} displays interpolation paths with five interpolants of four pairs of data from CIFAR10, created using the same architecture trained on different base distributions.
We pick the best-performing model on BPD on test data from the multiple training runs for each base distribution.
We visually compare a linear interpolation using a Gaussian base distribution against a spherical linear interpolation using a vMF base distribution with different concentration values.
Naturally, the images in the center show the difference and the effects resulting from the choice of base distribution and, hence, the interpolation procedure.

Overall, the linear interpolation with a Gaussian tends to show mainly darker objects on brighter background (almost black and white images) in the middle of the interpolation path.
This is not the case for the spherical interpolations using a vMF base distribution.
Specifically, in the second example showing dogs, the checkerboard background of the left endpoint smoothly fades out for the vMF ($\kappa=2d$) model while the Gaussian shows an almost white background.
A similar effect happens in the last pair of images, highlighting the weaknesses of a linear interpolation once again.
By contrast, the vMF models generate images where those effects are either less prominent or non-existent, suggesting a path that strictly follows the data manifold.
We provide more interpolation paths on CIFAR10 in the supplementary material.

Figure~\ref{fig:interpolation_fmnist} depicts interpolation paths with five interpolants on two pairs of data from Fashion-MNIST.
In both cases, the Gaussian model produces suboptimal images with visible color changes, which is not consistent with the endpoints.
Furthermore, there is visible deformation of the clothing items.

\begin{figure}[t]
\center
\includegraphics[width=\textwidth]{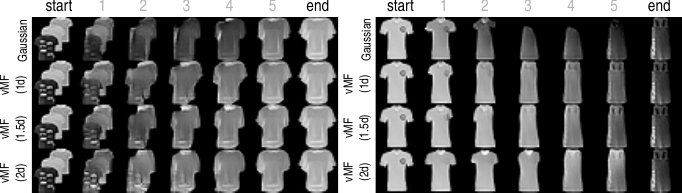}
\caption{Interpolation paths of two pairs from Fashion-MNIST using different models.}
\label{fig:interpolation_fmnist}
\end{figure}

\section{Related Work}\label{sec:related_work}

Interpolations are commonplace in generative modeling, being particularly useful for evaluating them.
Spherical linear interpolations \cite{shoemake1985animating} are also proposed \cite{white2016sampling} to circumvent the problems depicted in Figure~\ref{fig:intro} in GANs and VAEs.
However, as the Gaussian is kept as a base distribution, the difference in norms causes problems similar to the norm corrected approach.
The problem of interpolation is also investigated for GANs \cite{agustsson2019optimal}.
Specifically, they show that the quality of the generated images in the interpolation path improves when attempting to match the distribution of norms between interpolants and the GAN prior.
The problem with the distribution mismatch while interpolating is also studied in \cite{kilcher2018semantic}.

Simultaneously learning a manifold and corresponding normalizing flow on it is also possible \cite{brehmer2020flows}.
By contrast, in this paper, we employ a prescribed manifold, i.e., a $p$-norm sphere, on which the interpolation can be done in a principled way.
Using a vMF distribution as a prior of VAEs is also used to encourage the model to learn better latent representations on data with hyperspherical structure \cite{davidson2018hyperspherical,xu2018spherical}.
While results show improvements over a Gaussian prior, properties of our interest, such as interpolation, are not addressed.

Employing normalizing flows on non-Euclidean spaces, such as the hypersphere, was first proposed by \cite{gemici2016normalizing}.
They introduce a mapping for doing normalizing flows on hyperspherical data.
The main difference from our setting is that the data is already on a sphere and is moved to $\R^d$, an unrestricted space, performing the entire flow in there instead, before moving back to the sphere.
This avoids defining a flow on the sphere, which is studied in \cite{rezende2020normalizing} for tori and spheres.
Besides, normalizing flows on hyperbolic spaces are beneficial for graph-structured data~\cite{bose2020latent}.

A geometric analysis of autoencoders, showing that they learn latent spaces, which can be characterized by a Riemannian metric, is provided by \cite{arvanitidis2018latent}.
With this, interpolations follow a geodesic path under this metric, leading to higher quality interpolations.
Compared to our contribution, these approaches do not change the standard priors but propose alternative ways to interpolate samples.
In contrast, we propose an orthogonal approach by changing the base distribution and imposing constraints on the representation in our training procedure.
Consequently, standard interpolation procedures, such as the spherical linear interpolation, can be used in a principled way.

\section{Conclusion}\label{sec:conclusion}

This paper highlighted the limitations of linear interpolation in flow-based generative models using a Gaussian base distribution.
As a remedy, we proposed to focus on base representations with a fixed norm where the interpolation naturally overcomes those limitations and introduced normalizing flows onto unit $p$-norm spheres.
Specifically, we showed for the cases $p\in\{1,2\}$ that we could operate on the unit simplex and unit hypersphere, respectively.
We introduced a computationally efficient way of using a Dirichlet distribution as a base distribution for the case of $p=1$ and leveraged a von Mises-Fisher distribution using a stereographic projection onto a hypersphere for the case $p=2$.
Although the former suffered from numerical instabilities in a few experiments, our experimental results showed superior performance in terms of bits per dimension on test data and FID and KID scores on interpolation paths that resulted in natural transitions from one image to another.
This was also confirmed by visually comparing interpolation paths on CIFAR10 and Fashion-MNIST.

\section*{Acknowledgements}
This research was financed in part by the Coordena\c c\~ao de Aperfei\c coamento de Pessoal de N\'ivel Superior (CAPES), Brazil, Finance Code 001 and by FAPESP (grants 2017/24005-2 and 2018/19350-5).

\bibliographystyle{splncs04}
\bibliography{references}

\begin{thebibliography}{10}
\providecommand{\url}[1]{\texttt{#1}}
\providecommand{\urlprefix}{URL }
\providecommand{\doi}[1]{https://doi.org/#1}

\bibitem{agustsson2019optimal}
Agustsson, E., Sage, A., Timofte, R., Gool, L.V.: Optimal transport maps for
  distribution preserving operations on latent spaces of generative models. In:
  International Conference on Learning Representations (2019)

\bibitem{arvanitidis2018latent}
Arvanitidis, G., Hansen, L.K., Hauberg, S.: Latent space oddity: on the
  curvature of deep generative models. In: International Conference on Learning
  Representations (2018)

\bibitem{binkowski2018demystifying}
Bińkowski, M., Sutherland, D.J., Arbel, M., Gretton, A.: Demystifying {MMD}
  {GAN}s. In: International Conference on Learning Representations (2018)

\bibitem{bose2020latent}
Bose, J., Smofsky, A., Liao, R., Panangaden, P., Hamilton, W.: Latent variable
  modelling with hyperbolic normalizing flows. In: International Conference on
  Machine Learning. pp. 1045--1055. PMLR (2020)

\bibitem{brehmer2020flows}
Brehmer, J., Cranmer, K.: Flows for simultaneous manifold learning and density
  estimation. In: Advances in Neural Information Processing Systems. vol.~33
  (2020)

\bibitem{chen2019residual}
Chen, T.Q., Behrmann, J., Duvenaud, D.K., Jacobsen, J.H.: Residual flows for
  invertible generative modeling. In: Advances in Neural Information Processing
  Systems. pp. 9913--9923 (2019)

\bibitem{clanuwat2018deep}
Clanuwat, T., Bober-Irizar, M., Kitamoto, A., Lamb, A., Yamamoto, K., Ha, D.:
  Deep learning for classical {J}apanese literature. CoRR
  \textbf{abs/1812.01718} (2018)

\bibitem{davidson2018hyperspherical}
Davidson, T.R., Falorsi, L., De~Cao, N., Kipf, T., Tomczak, J.M.:
  Hyperspherical variational auto-encoders. In: Proceedings of the
  Thirty-Fourth Conference on Uncertainty in Artificial Intelligence. pp.
  856--865 (2018)

\bibitem{dinh2014nice}
Dinh, L., Krueger, D., Bengio, Y.: {NICE:} non-linear independent components
  estimation. In: International Conference on Learning Representations (2015)

\bibitem{dinh2016density}
Dinh, L., Shol-Dickstein, J., Bengio, S.: {Density estimation using Real NVP}.
  In: International Conference on Learning Representations (2017)

\bibitem{gemici2016normalizing}
Gemici, M.C., Rezende, D., Mohamed, S.: Normalizing flows on {R}iemannian
  manifolds. CoRR  \textbf{abs/1611.02304} (2016)

\bibitem{goodfellow2014generative}
Goodfellow, I., Pouget-Abadie, J., Mirza, M., Xu, B., Warde-Farley, D., Ozair,
  S., Courville, A., Bengio, Y.: Generative adversarial nets. In: Advances in
  Neural Information Processing Systems. pp. 2672--2680 (2014)

\bibitem{heusel2017gans}
Heusel, M., Ramsauer, H., Unterthiner, T., Nessler, B., Hochreiter, S.: {GANs}
  trained by a two time-scale update rule converge to a local nash equilibrium.
  In: Advances in Neural Information Processing Systems. pp. 6626--6637 (2017)

\bibitem{karras2018progressive}
Karras, T., Aila, T., Laine, S., Lehtinen, J.: Progressive growing of {GAN}s
  for improved quality, stability, and variation. In: International Conference
  on Learning Representations (2018)

\bibitem{kilcher2018semantic}
Kilcher, Y., Lucchi, A., Hofmann, T.: Semantic interpolation in implicit
  models. In: International Conference on Learning Representations (2018)

\bibitem{kingma2014adam}
Kingma, D.P., Ba, J.: Adam: {A} method for stochastic optimization. In:
  International Conference on Learning Representations (2015)

\bibitem{kingma2013auto}
Kingma, D.P., Welling, M.: Auto-encoding variational {B}ayes. In: International
  Conference on Learning Representations (2014)

\bibitem{kingma2019introduction}
Kingma, D.P., Welling, M., et~al.: An introduction to variational autoencoders.
  Foundations and Trends{\textregistered} in Machine Learning  \textbf{12}(4),
  307--392 (2019)

\bibitem{kingma2018glow}
Kingma, D.P., Dhariwal, P.: Glow: Generative flow with invertible 1x1
  convolutions. In: Advances in Neural Information Processing Systems. pp.
  10215--10224 (2018)

\bibitem{kingma2016improved}
Kingma, D.P., Salimans, T., Jozefowicz, R., Chen, X., Sutskever, I., Welling,
  M.: Improved variational inference with inverse autoregressive flow. In:
  Advances in Neural Information Processing Systems. pp. 4743--4751 (2016)

\bibitem{krizhevsky2009learning}
Krizhevsky, A., Hinton, G.: Learning multiple layers of features from tiny
  images. Tech. rep. (2009)

\bibitem{lecun2010mnist}
LeCun, Y., Cortes, C., Burges, C.: Mnist handwritten digit database. ATT Labs
  [Online]. Available: http://yann.lecun.com/exdb/mnist  \textbf{2} (2010)

\bibitem{papamakarios2019normalizing}
Papamakarios, G., Nalisnick, E., Rezende, D.J., Mohamed, S., Lakshminarayanan,
  B.: Normalizing flows for probabilistic modeling and inference. Journal of
  Machine Learning Research  \textbf{22}(57),  1--64 (2021)

\bibitem{PyTorch2019}
Paszke, A., Gross, S., Massa, F., Lerer, A., Bradbury, J., Chanan, G., Killeen,
  T., Lin, Z., Gimelshein, N., Antiga, L., Desmaison, A., Kopf, A., Yang, E.,
  DeVito, Z., Raison, M., Tejani, A., Chilamkurthy, S., Steiner, B., Fang, L.,
  Bai, J., Chintala, S.: Pytorch: An imperative style, high-performance deep
  learning library. In: Advances in Neural Information Processing Systems, pp.
  8024--8035 (2019)

\bibitem{radford2015unsupervised}
Radford, A., Metz, L., Chintala, S.: Unsupervised representation learning with
  deep convolutional generative adversarial networks. In: International
  Conference on Learning Representations (2016)

\bibitem{rezende2015variational}
Rezende, D., Mohamed, S.: Variational inference with normalizing flows. In:
  International Conference on Machine Learning. pp. 1530--1538. PMLR (2015)

\bibitem{rezende2020normalizing}
Rezende, D.J., Papamakarios, G., Racaniere, S., Albergo, M., Kanwar, G.,
  Shanahan, P., Cranmer, K.: Normalizing flows on tori and spheres. In:
  International Conference on Machine Learning. pp. 8083--8092 (2020)

\bibitem{rippel2013high}
Rippel, O., Adams, R.P.: High-dimensional probability estimation with deep
  density models. CoRR  \textbf{abs/1302.5125} (2013)

\bibitem{salimans2016improved}
Salimans, T., Goodfellow, I., Zaremba, W., Cheung, V., Radford, A., Chen, X.:
  Improved techniques for training {GAN}s. In: Advances in Neural Information
  Processing Systems. pp. 2234--2242 (2016)

\bibitem{shoemake1985animating}
Shoemake, K.: Animating rotation with quaternion curves. In: Proceedings of the
  12th Annual Conference on Computer Graphics and Interactive Techniques. pp.
  245--254 (1985)

\bibitem{tabak2013family}
Tabak, E.G., Turner, C.V.: A family of nonparametric density estimation
  algorithms. Communications on Pure and Applied Mathematics  \textbf{66}(2),
  145--164 (2013)

\bibitem{tabak2010density}
Tabak, E.G., Vanden-Eijnden, E.: Density estimation by dual ascent of the
  log-likelihood. Communications in Mathematical Sciences  \textbf{8}(1),
  217--233 (2010)

\bibitem{white2016sampling}
White, T.: Sampling generative networks. CoRR  \textbf{abs/1609.04468} (2016)

\bibitem{xiao2017fashion}
Xiao, H., Rasul, K., Vollgraf, R.: Fashion-mnist: a novel image dataset for
  benchmarking machine learning algorithms. CoRR  \textbf{abs/1708.07747}
  (2017)

\bibitem{xu2018spherical}
Xu, J., Durrett, G.: Spherical latent spaces for stable variational
  autoencoders. In: Proceedings of the 2018 Conference on Empirical Methods in
  Natural Language Processing. pp. 4503--4513 (2018)

\end{thebibliography}

\appendix

\begin{figure}[h!]
\center
\includegraphics[width=\textwidth]{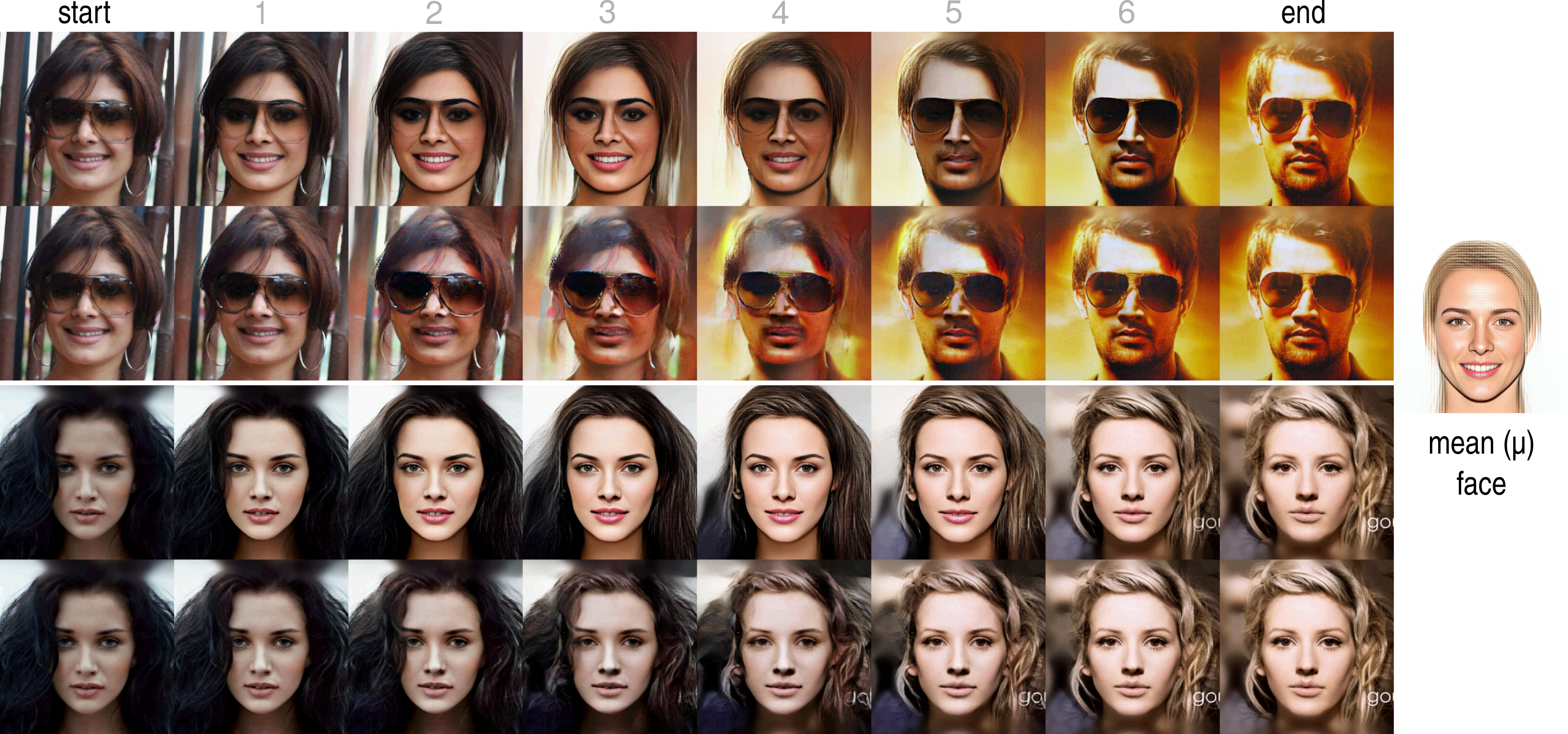}
\caption{Left: Linear (1st and 3rd row) and norm-corrected (2nd and 4th row) interpolations paths. Right: decoded expectation of base distribution, i.e., the mean face.}
\label{fig:interpol2}
\end{figure}

\section{More Interpolation Paths on CelebA}

Figure~\ref{fig:interpol2} depicts two additional interpolation paths of images taken from CelebA.
Analogously to Figure~\ref{fig:interpol1}, the top row shows a linear interpolation and the bottom row an interpolation with norm correction as introduced in Equation~\eqref{eq-heuristic}.
While this correction guarantees that the norms of interpolants stay within the observed range of data, we note that this is a rather ad-hoc way to perform the interpolation as we will point out in the remainder. 

The leftmost and rightmost faces in Figure~\ref{fig:interpol2} are real data while the remaining ones are interpolants.
The mean face shown on the right hand side is clearly visible in the central interpolants; e.g., the glasses disappear in the top row around the center and then reappear while the lips in the third row become more prominent towards the center.
We credit both to the properties of the mean face.
In contrast, the norm-corrected interpolation does not suffer from distortions by the mean face and has many desirable properties, but also a major limitation.
The first and last interpolants are very close to the end points, which are real data.
Figure~\ref{fig:interpolation_nclerp} depicts the same phenomenon for CIFAR10.

\begin{table}[h!]
\center
\caption{
Results for interpolation averaged over three independent runs including standard errors.
Interpolations are within and across classes and use five intermediate points;
\emph{lerp} refers to a linear interpolation; \emph{nclerp} refers to
the norm-corrected linear interpolation (Section~\ref{sec:heuristic}) and
\emph{slerp} refers to the spherical interpolation.
}
\label{tab:results2}
\begin{tabular}{rcrcccc}
\toprule
& Base dist. & Type & BPD & FID & KID \\
\midrule 
\multirow{6}{*}{\rotatebox[origin=c]{90}{MNIST}}
 & Gaussian               &   lerp & \textbf{1.34} \tiny{$\pm$   0.06} &         9.78  \tiny{$\pm$   0.50} &         0.007  \tiny{$\pm$  0.001} \\
 & Gaussian               & nclerp &         1.50  \tiny{$\pm$   0.12} &        11.66  \tiny{$\pm$   1.06} &         0.009  \tiny{$\pm$  0.001} \\
 & vMF $\kappa=1d$        &  slerp & \textbf{1.34} \tiny{$\pm$   0.10} &         6.70  \tiny{$\pm$   0.46} &         0.005  \tiny{$\pm$  0.001} \\
 & vMF $\kappa=1.5d$      &  slerp &         1.41  \tiny{$\pm$   0.09} & \textbf{6.65} \tiny{$\pm$   0.20} & \textbf{0.004} \tiny{$\pm$  0.000} \\
 & vMF $\kappa=2d$        &  slerp &         1.65  \tiny{$\pm$   0.09} &         8.31  \tiny{$\pm$   0.20} &         0.006  \tiny{$\pm$  0.000} \\
 & Dirichlet $\alpha=2$   &   lerp &         1.61  \tiny{$\pm$   0.10} &         8.96  \tiny{$\pm$   0.32} &         0.006  \tiny{$\pm$  0.000} \\
\midrule 
\multirow{6}{*}{\rotatebox[origin=c]{90}{K-MNIST}}
 & Gaussian               &   lerp &         2.00  \tiny{$\pm$   0.16} &          31.51  \tiny{$\pm$   2.53} &         0.034  \tiny{$\pm$  0.003} \\
 & Gaussian               & nclerp &         1.85  \tiny{$\pm$   0.18} &          29.76  \tiny{$\pm$   2.44} &         0.033  \tiny{$\pm$  0.003} \\
 & vMF $\kappa=1d$        &  slerp &         2.15  \tiny{$\pm$   0.13} &          29.39  \tiny{$\pm$   5.66} &         0.032  \tiny{$\pm$  0.006} \\
 & vMF $\kappa=1.5d$      &  slerp & \textbf{1.65} \tiny{$\pm$   0.06} &          37.35  \tiny{$\pm$   3.49} &         0.041  \tiny{$\pm$  0.004} \\
 & vMF $\kappa=2d$        &  slerp &         2.09  \tiny{$\pm$   0.13} &  \textbf{23.74} \tiny{$\pm$   4.41} & \textbf{0.026} \tiny{$\pm$  0.005} \\
 & Dirichlet $\alpha=2$   &   lerp &         1.90  \tiny{$\pm$   0.04} &          37.80  \tiny{$\pm$   4.30} &         0.042  \tiny{$\pm$  0.005} \\
\midrule 
\multirow{5}{*}{\rotatebox[origin=c]{90}{F-MNIST}}
 & Gaussian               &   lerp &         2.84  \tiny{$\pm$   0.04} &  \textbf{16.93} \tiny{$\pm$   0.06} &         0.011  \tiny{$\pm$  0.000} \\
 & Gaussian               & nclerp &         2.86  \tiny{$\pm$   0.03} &  \textit{13.61} \tiny{$\pm$   0.52} & \textbf{0.009} \tiny{$\pm$  0.000} \\
 & vMF $\kappa=1d$        &  slerp & \textbf{2.69} \tiny{$\pm$   0.02} &          20.10  \tiny{$\pm$   2.11} &         0.012  \tiny{$\pm$  0.001} \\
 & vMF $\kappa=1.5d$      &  slerp &         2.78  \tiny{$\pm$   0.05} &          17.56  \tiny{$\pm$   1.46} & \textbf{0.009} \tiny{$\pm$  0.001} \\
 & vMF $\kappa=2d$        &  slerp &         2.72  \tiny{$\pm$   0.06} &          21.41  \tiny{$\pm$   0.79} &         0.013  \tiny{$\pm$  0.001} \\
\midrule 
\multirow{5}{*}{\rotatebox[origin=c]{90}{CIFAR10}}
 & Gaussian               &   lerp &          2.82  \tiny{$\pm$   0.04} &          63.17  \tiny{$\pm$   0.99} &         0.059  \tiny{$\pm$  0.001} \\
 & Gaussian               & nclerp &          3.33  \tiny{$\pm$   0.01} &  \textit{16.83} \tiny{$\pm$   0.24} & \textit{0.013} \tiny{$\pm$  0.000} \\
 & vMF $\kappa=1d$        &  slerp &          2.93  \tiny{$\pm$   0.04} &  \textbf{56.13} \tiny{$\pm$   0.38} & \textbf{0.049} \tiny{$\pm$  0.000} \\
 & vMF $\kappa=1.5d$      &  slerp &          2.85  \tiny{$\pm$   0.04} &          59.89  \tiny{$\pm$   4.95} &         0.054  \tiny{$\pm$  0.004} \\
 & vMF $\kappa=2d$        &  slerp &  \textbf{2.79} \tiny{$\pm$   0.06} &          57.14  \tiny{$\pm$   3.45} &         0.051  \tiny{$\pm$  0.004} \\
\bottomrule
\end{tabular}
\end{table}

\section{Interpolations both across and within classes}

Table~\ref{tab:results2} depicts the same experiments as shown in Table~\ref{tab:results_interpolation}.
The difference, however, is that the interpolations are no longer restricted to be within classes but uniformly sampled and thus also contain interpolations across classes.
Using the vMF as a base distribution clearly outperforms the Gaussian in terms of bits per dimension on test data.
The only exception is a tie on MNIST.
This is caused by the same effect that caused Figure~\ref{fig:intro}: 
since the norm drops when conducting a linear interpolation, the interpolants are much closer to the mean.
Hence, they possess higher likelihoods in terms of the base distribution and yield better scores, because bits per dimension is a rescaled negative log-likelihood.
The results in terms of FID and KID scores on general interpolations are in line with the results in Table~\ref{tab:results_interpolation}: the vMF base distribution yields better scores on MNIST and K-MNIST.
The norm-corrected linear interpolation (\emph{nclerp}) yields lower FID and KID scores on K-MNIST and CIFAR10.
We credit this to the biased evaluation as discussed in Section~\ref{sec:heuristic}.

\section{Samples from the models}

\begin{figure}[t]
\center
\includegraphics[width=.49\textwidth]{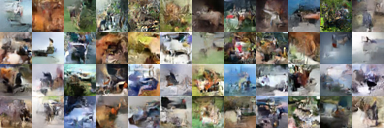}
\includegraphics[width=.49\textwidth]{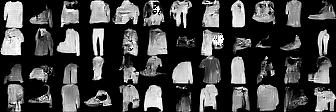}
\includegraphics[width=.49\textwidth]{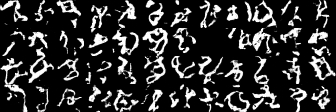}
\includegraphics[width=.49\textwidth]{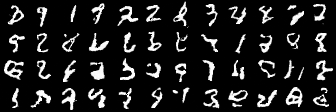}
\caption{Samples generated from the models for CIFAR10 (first block), Fashion-MNIST (second block), Kuzushiji-MNIST (third block), and MNIST (fourth block). Per data set, we show one row of samples from Gaussian, vMF ($\kappa=1d$), vMF ($\kappa=1.5d$), and vMF ($\kappa=2d$), where the order is top to bottom.}
\label{fig:samples}
\end{figure}

Figure~\ref{fig:samples} shows twelve samples per model.
The figure essentially shows that the choice of the base distribution does not influence the quality of the generated samples.
Changing the base distribution thus leads to more natural interpolations without sacrificing generative performance.
Note that better looking samples can be produced with a more sophisticated architecture, but this is not the goal of this paper.

\section{More interpolations on CIFAR10}

\begin{figure}[t]
\center
\includegraphics[width=\textwidth]{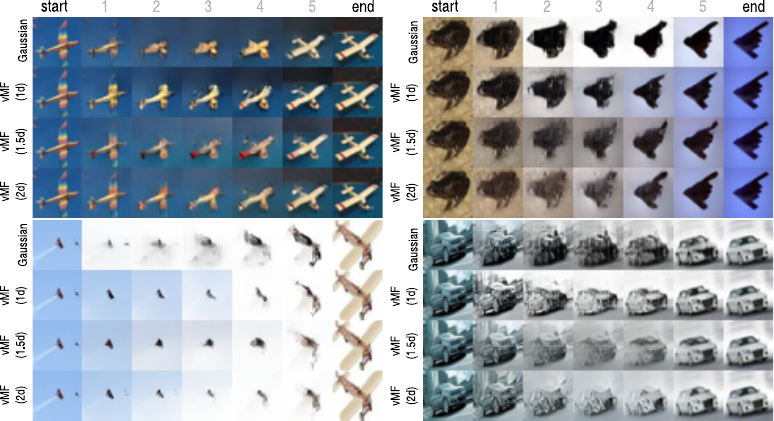}
\caption{Interpolation paths of four pairs of data from \mbox{CIFAR10} using different models.
}
\label{fig:interpolation_cifar10_extra}
\end{figure}

Figure~\ref{fig:interpolation_cifar10_extra} shows interpolation paths on four additional pairs of images from CIFAR10.
The order is the same as in Figure~\ref{fig:interpolation_cifar10}.
The first row shows the Gaussian, the second, third, and fourth rows depict the vMF with $\kappa \in \{ 1d, 1.5d, 2d \}$, respectively.
Within the top left part, the Gaussian prior yields the worst interpolation path since the airplane loses its wings.
The vMF ($\kappa=1d$) shows the most natural transition.
While a meaningful transition of the images in the top right and bottom left parts is not obvious, the Gaussian base distribution fails to produce a smooth transition of the background and falls back to white.
In contrast, all vMF models yield a smooth change of background.
The interpolation of the cars depicted in the bottom right part shows an example where both the Gaussian as well as the vMF ($\kappa=1.5d$) have issues.

\end{document}